\documentclass{article}

\usepackage[preprint]{timeseries_workshop}

\usepackage[utf8]{inputenc} 
\usepackage[T1]{fontenc}    
\usepackage{hyperref}       
\usepackage{url}            
\usepackage{booktabs}       
\usepackage{amsfonts}       
\usepackage{nicefrac}       
\usepackage{microtype}      
\usepackage{xcolor}         

\usepackage{graphicx}
\usepackage{svg}

\title{There is No "apple" in Timeseries:\\ Rethinking TSFM through the Lens of Invariance}

\author{%
    Arian Prabowo\thanks{\url{www.arianprabowo.com}}\\
    School of Computer Science and Engineering\\
    University of New South Wales\\
    Kensington, NSW 2052\\
    \texttt{arian.prabowo@unsw.edu.au}
    \AND
    Flora D. Salim\\
    School of Computer Science and Engineering\\
    University of New South Wales\\
    Kensington, NSW 2052\\
    \texttt{flora.salim@unsw.edu.au}
}

\begin{document}

\maketitle

\begin{abstract}
Timeseries foundation models (TSFMs) have multiplied, yet lightweight supervised baselines and even classical models often match them.
We argue this gap stems from the naïve importation of NLP/CV pipelines.
In language and vision, large web-scale corpora densely capture human concepts i.e. there are countless images and text of apples.
In contrast, timeseries data is built to complement the image and text modalities.
There are no timeseries dataset that contains the concept "apple".
As a result, the “scrape everything online” paradigm fails for TS.
We posit that progress demands a shift from opportunistic aggregation to principled design: constructing datasets that systematically span the space of invariance that preserve temporal semantics.
To this end, we suggest that the ontology of timeseries invariances should be built based on first principles.
Only by ensuring representational completeness through invariance coverage can TSFMs achieve the aligned structure necessary for generalisation, reasoning, and truly emergent behaviour.
\end{abstract}

\section{The Incompleteness of Timeseries Data}

\begin{figure}[htbp]
  \centering
  \includegraphics[width=.8\textwidth]{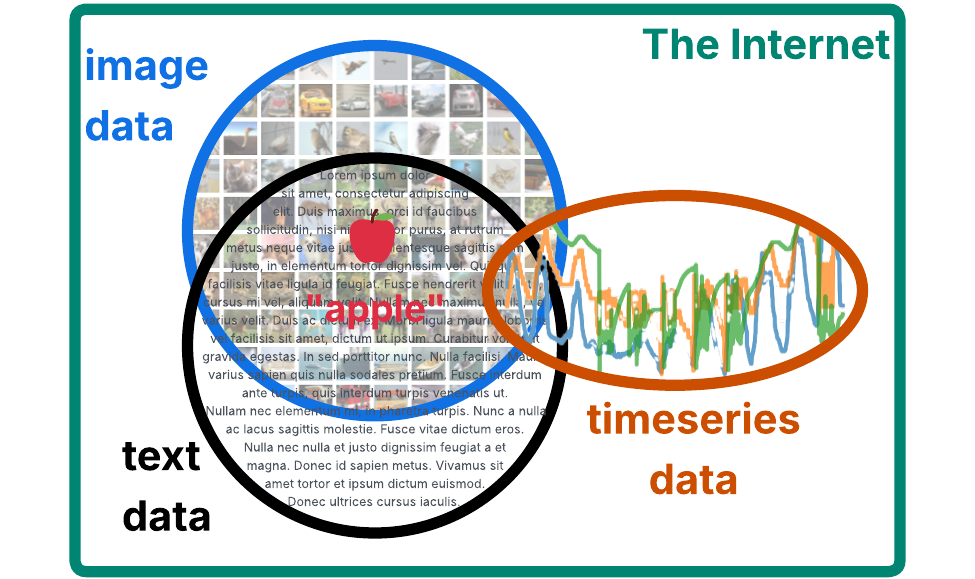}
  \caption{
   The “scrape everything on the internet” paradigm succeeds in vision and language because web-scale image and text corpora collectively span most of the human-relevant representational space.
   In contrast, timeseries data occupy a complementary subspace—capturing dynamics, rhythms, and control signals that rarely overlap with such semantic concepts.  
   The slice of reality containing the notion of “apple”~\cite{huh2024PRH}, for example, is densely represented in image and text corpora but absent from timeseries.
   We picked the "apple" example in direct reference to the first equation in the \href{https://phillipi.github.io/prh/}{Platonic Represntation Hypothesis (PRH) webpage}.
   }
    \label{fig:vizabs}
\end{figure}

\textbf{Foundation models (FM) are trained on “scrape everything (on the internet)” corpus.}
This has been proven effective in the domain of computer vision (CV) and natural language processing (NLP).
Timeseries foundation models (TSFMs) are gaining popularity by following this paradigm.
However, this has recently been questioned due to its poor performance over a fully supervised baseline, hence the missing BERT moment~\cite{xu2025specialized}.
This paper argues that the “scrape everything” approach is effective for text and image data but fundamentally unsuitable for time-series.

\textbf{Image and text data on the internet are near human-representationally complete.}
This stems from bias in data collection due to human behaviour. 
Humans naturally document the world through writing, photography, and video.
As a result, the internet is saturated with multimodal data that forms an almost complete, human-relevant representation of the world.
This completeness is not merely an aspect of scale but of semantic coverage: the web collectively encodes nearly every concept, object, and event that humans find meaningful.
Foundation models trained on such data therefore perform well on a wide range of human-relevant tasks~\cite{awais2025foundation}.
Fig.~\ref{fig:vizabs} shows that the internet is dominated by image and text data.

\textbf{The Platonic Representation Hypothesis (PRH)~\cite{huh2024PRH} provides supporting evidence for the completeness of the image and text corpus.}
The PRH posits that “neural networks, trained with different objectives on different data and modalities, are converging to a shared statistical model of reality in their representation spaces.”
Such convergence is only possible if each pretraining dataset independently provides a sufficiently complete capture of reality.
Fig.~\ref{fig:vizabs} shows a huge intersection between the image and text data.

\textbf{However, timeseries data are not human-representationally complete.}
They are collected precisely to capture aspects of reality that images and text cannot, serving as complementary rather than overlapping modalities.
Borrowing the example in the original PRH paper, the concept of “apple” is richly represented in both text and images through abundant descriptions and visuals, yet no time-series signal conveys its meaning:
\textbf{there is no "apple" in timeseries}.
As shown in Fig.~\ref{fig:vizabs}, “apple” lies within the intersection of image and text, but outside time-series data.
The reverse also applies; many time-series phenomena cannot be effectively expressed through language.
Road traffic, for example, is typically described only in broad qualitative categories such as free-flowing, congested, or peak/off-peak.
This complementary configuration explains why the “scrape-everything” paradigm succeeds for text and images but fails for timeseries.
Even with perfect coverage and training, a timeseries foundation model would lack emergent common-sense or zero-shot behaviours, since the semantic and physical regularities underpinning real-world understanding are absent from timeseries measurements.

\textbf{Existing TSFM pretraining datasets are incomplete.}
Because human-collected timeseries data only capture a narrow slice of reality, this incompleteness inevitably carries over into the TSFM pretraining datasets derived from them.
These datasets are assembled through opportunistic and ad hoc processes, guided by data availability rather than by a systematic effort to achieve representational completeness of the real world.
Some draw from previously established collections such as the
Time Series Pile~\cite{goswami2024moment},
LOTSA~\cite{woo2024uni2ts}, and
UTSD~\cite{liu2024timer},
which themselves aggregate prior benchmarks like the
Monash Forecasting Archive~\cite{godahewa2021monash} and 
Informer datasets~\cite{zhou2021informer}.
Others augment public data with in-house sources.
For instance,
TimesFM~\cite{das2024timesFM} relies on Google Trends,
TimeHF~\cite{qi2025timeHF} incorporates JD supply-chain sales covering over 20 000 products,and 
BOOM~\cite{cohen2025toto} pretrains on telemetry metrics from DataDog’s cloud systems.
In addition, several models expand coverage through synthetic generation:
TimeHF applies data augmentation,
Chronos~\cite{ansari2024chronos} and WaveToken~\cite{masserano2025wavetoken} introduce Gaussian process–based synthetic series,
and BOOM uses procedurally generated rule-based signals.

\textbf{Efforts to mitigate this incompleteness remain superficial.}
The Time Series Pile seeks broader domain coverage through aggregation rather than systematic design.
MOIRAI enforces a 0.1\% per-dataset cap within LOTSA—an ad hoc constraint that limits dominance without addressing underlying representational bias.
TimeHF supplements its retail-focused corpus with public weather and energy datasets.
BOOM integrates open benchmarks with telemetry logs, incrementally increasing domain variance.
None of these approaches constitute a systematic effort toward representational completeness.
As a result, TSFMs continue to underperform in zero-shot settings, lacking the emergent reasoning and world-aligned understanding that arise only from exposure to semantically complete data.


The implication of these observations are is that TSFM models require a data paradigm fundamentally distinct from those of vision and language.
Progress will not come from scaling alone but from constructing datasets that systematically span the diversity of physical and behavioural processes underlying the world’s dynamics.
Identifying the combination of timeseries domains that together approximate a world-complete representation of reality remains an open research question.
In the following section, we argue that \textbf{the lens of invariance provides a principled framework for pursuing such completeness}.

\section{Timeseries Invariance Ontology}

\begin{figure}[htbp]
    \centering
    \includegraphics[width=.7\textwidth]{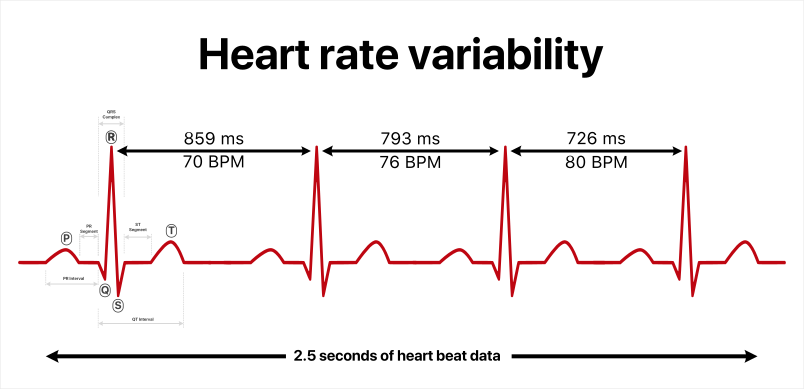} 
    \includegraphics[width=1\textwidth]{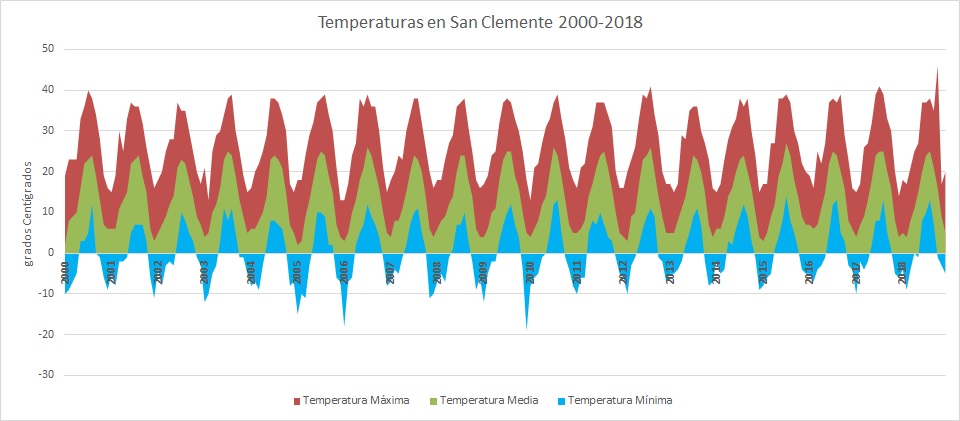}
    \includegraphics[width=.5\textwidth]{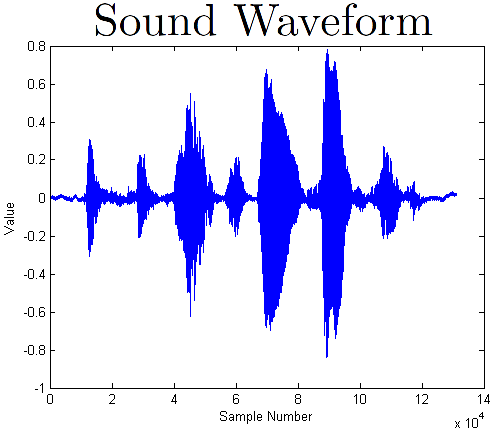}
    \caption{
    Illustration of invariances in various domains in timeseries.
    Minor temporal shifts preserve meaning in ECG signals through heart rate variability, yet the same shifts in meteorological data misalign seasonal cycles.
    Similarly, vertical inversion leaves audio waveforms perceptually unchanged, but in ECG it indicates lead reversal, and in climate data it represents a physically invalid temperature inversion.
    (Images adapted from publicly available sources.)
    }
    \label{fig:invariance_eg}
\end{figure}

A defining property of timeseries domains is the presence of domain-specific invariances and equivariances.
These are transformations that preserve a signal’s underlying dynamics, or at least should not alter its semantic or predictive identity~\cite{bronstein2021GDL}.
Identifying and leveraging these invariances is essential for both model design and dataset construction.  
Different timeseries domains exhibit distinct invariance structures, yet existing benchmarks and pretraining corpora neither enumerate nor balance them.  
As a result, current models risk overfitting to superficial correlations rather than learning robust representations that generalise across domains.  
This section outlines several illustrative examples.

\textbf{Spectral Invariance.}  
In high-frequency signals such as audio, vibration, and speech waveforms, information is primarily encoded in spectral composition.  
These domains exhibit spectral invariance, where small shifts or scalings in frequency content do not alter the underlying semantics.  
For instance, the spoken word “hello” remains identifiable despite differences in pitch or volume.  
When a pretraining corpus is dominated by slow-varying, seasonal domains such as economics or climate data, models may never encounter such invariance classes, limiting cross-modal generalisation and the emergence of temporal reasoning.
Conversely, models that internalise spectral invariance too broadly may misinterpret meaningful frequency shifts in domains where they encode genuine structural change—for example, economic cycles, market volatility, or hydrological regime shifts.

\textbf{Amplitude Invariance.} 
In many sensing contexts, scaling intensity or measurement units does not change the underlying behaviour.
Devices such as microphones, accelerometers, or pressure sensors often have adjustable or drifting gain, making relative changes more informative than absolute values.
However, amplitude invariance is not universal.
In domains such as energy load, traffic, or physiology, amplitude itself carries semantic meaning—signalling demand surges, congestion, or heartbeat anomalies.
Pretraining on corpora that disregard amplitude sensitivity may therefore erase meaningful information, while training exclusively on amplitude-sensitive domains can have the opposite effect, producing models that fail to recognise shape- or pattern-level equivalences across sensors.
Balanced exposure to both invariance regimes is necessary for timeseries foundation models to generalise across heterogeneous sensing conditions.

\textbf{Shape (Morphological) Invariance.}  
In certain domains, local waveform shape rather than absolute duration or amplitude defines identity.  
This property is characteristic of ECG traces, gesture dynamics, and speech signals, where transient motifs at multiple scales convey meaning.  
By contrast, domains such as climate or retail sales rely more on aggregate magnitudes.  
Models that fail to capture morphological invariance struggle to recognise recurring motifs that vary in scale or duration.

\textbf{Elastic Morphological Invariance.}
In certain domains, the local waveform shape, rather than absolute duration or amplitude,defines identity.
Signals such as ECG traces, gestures, and trajectories share this property: the same underlying pattern may occur at different speeds or with local temporal distortions.
This reflects not only morphological invariance but also temporal warping invariance, where nonlinear time deformations preserve semantic identity.
Dynamic Time Warping (DTW) and related alignment methods embody this assumption, enabling comparison of signals that differ in rate or alignment.
By contrast, domains such as climate, finance, or control systems encode meaning in the precise timing of events, where warping destroys causal structure.
Models that fail to capture these dual invariances struggle to recognise recurring motifs under variable durations, while those that overgeneralise time-warping risk erasing meaningful chronological information.

\textbf{Distributional (Stochastic) Invariance.}  
A distinct form of invariance arises in stochastic domains, where the semantic identity of a process is defined not by an individual trajectory but by its underlying probability law.  
Temporal permutations or alternative realisations preserve process identity: white noise, Brownian motion, or AR(1) dynamics remain equivalent under resampling.  
Such invariance characterises domains like turbulence, diffusion, random walks, and financial returns, where statistical moments and spectral densities define behaviour.  
Deterministic domains such as ECG or traffic lack this property, as the specific sequence of events carries meaning.

\textbf{Parametric Invariance.}  
In many dynamical systems, the governing parameters remain stable within local regimes, such as viscosity in fluid flow, friction coefficients in mechanics, or reaction rates in chemistry.  
A model capable of inferring these latent constants demonstrates deeper understanding of the underlying process, effectively reconstructing its governing equations from data.  
Domains like fluid dynamics, epidemiology, and controlled mechanical systems exhibit strong parametric invariance, making them amenable to structured pretraining.  
Conversely, domains subject to frequent regime changes (e.g. speech, finance, and traffic) lack persistent constants and exhibit piecewise or context-dependent dynamics.
Models overfitted to constant-regime data may therefore be fragile to transient changes in such domains.

\textbf{Toward a Timeseries Invariance Ontology.}
The diversity of invariances across timeseries domains underscores that there is no single symmetry governing temporal data.
To move beyond opportunistic dataset aggregation, the community must develop a formal ontology of timeseries invariances.
Such an ontology would not only guide model architectures and evaluation but also inform how data should be collected, balanced, or even synthetically generated to ensure comprehensive coverage of invariance space.
A world-complete timeseries corpus must therefore be designed, not merely scraped—curated to represent the full spectrum of dynamical symmetries that define temporal reality.

\textbf{An ontology of timeseries invariance from first principles} could be a good starting point.
Formally, any invariance can be expressed as the set of transformations 
$g$ such that 
$T(g \cdot x)=T(x)$
, where
$T$ is a downstream task or representation operator.
From this foundation, one can systematically generate invariance candidates from a canonical set of primitive transformations on timeseries:
time-axis operations (shift, scaling, reversal, monotone warping);
value-space operations (affine, monotone, diffeomorphic, or stochastic channels);
index-space operations (permutations, isometries, graph automorphisms, or topology-preserving maps);
and dynamical or causal symmetries (structural conjugacy, conservation, symplectic, or causal-mechanism invariance).
Closing these transformation sets under composition yields a complete invariance group for temporal data—an explicit ontology that can directly guide both dataset curation and model design.

\section{Conclusion}

TSFMs underperform not only because of limited scale or model capacity, but because timeseries corpora are structurally incomplete.
They lack the human-semantic coverage that makes web-scale text and image datasets so powerful; i.e. there is no "apple" in timeseries.
The “scrape everything” strategy, successful in language and vision, cannot yield emergent understanding or robust zero-shot reasoning in the temporal domain.
Progress therefore requires a shift from ad hoc aggregation to deliberate design: datasets must be curated (and, where appropriate, synthetically generated) to systematically span the space defined by an ontology of timeseries invariances.
Constructing this ontology from first principles ensures completeness, grounding it in the full range of transformations that preserve temporal semantics.
Only under this paradigm can TSFMs develop the aligned structure necessary for generalisation, reasoning, and truly emergent behaviour.

\begin{ack}
This work makes use of the the following images from Wikimedia Commons:

\begin{itemize}
  \item \emph{"Heart rate variability (HRV)"} by David S. Goodsell, available at  
    \href{https://commons.wikimedia.org/wiki/File:Heart_rate_variability_(HRV).svg}{commons.wikimedia.org/wiki/File:Heart\_rate\_variability\_(HRV).svg},  
    licensed under the \href{https://creativecommons.org/licenses/by-sa/4.0/}{CC BY-SA 4.0} licence.

  \item \emph{"San Clemente. Temperaturas"} by Mtorrecilla, available at  
    \href{https://commons.wikimedia.org/wiki/File:San_Clemente._Temperaturas.jpg}{commons.wikimedia.org/wiki/File:San\_Clemente.\_Temperaturas.jpg},  
    licensed under the \href{https://creativecommons.org/licenses/by-sa/3.0/}{CC BY-SA 3.0} licence.

  \item \emph{"Haar DWT of the Sound Waveform “I Love Wavelets”"} available at  
    \href{https://en.wikipedia.org/wiki/File:Haar_DWT_of_the_Sound_Waveform_\%22I_Love_Wavelets\%22.png}{en.wikipedia.org/wiki/File:Haar\_DWT\_of\_the\_Sound\_Waveform\_"I\_Love\_Wavelets".png},  
    licensed under the \href{https://creativecommons.org/licenses/by-sa/3.0/}{CC BY-SA 3.0} licence.
\end{itemize}


\end{ack}

\bibliographystyle{plainnat}
\bibliography{1bib}

\begin{thebibliography}{14}
\providecommand{\natexlab}[1]{#1}
\providecommand{\url}[1]{\texttt{#1}}
\expandafter\ifx\csname urlstyle\endcsname\relax
  \providecommand{\doi}[1]{doi: #1}\else
  \providecommand{\doi}{doi: \begingroup \urlstyle{rm}\Url}\fi

\bibitem[Ansari et~al.(2024)Ansari, Stella, Turkmen, Zhang, Mercado, Shen, Shchur, Rangapuram, Pineda~Arango, Kapoor, Zschiegner, Maddix, Mahoney, Torkkola, Gordon~Wilson, Bohlke-Schneider, and Wang]{ansari2024chronos}
Abdul~Fatir Ansari, Lorenzo Stella, Caner Turkmen, Xiyuan Zhang, Pedro Mercado, Huibin Shen, Oleksandr Shchur, Syama~Syndar Rangapuram, Sebastian Pineda~Arango, Shubham Kapoor, Jasper Zschiegner, Danielle~C. Maddix, Michael~W. Mahoney, Kari Torkkola, Andrew Gordon~Wilson, Michael Bohlke-Schneider, and Yuyang Wang.
\newblock Chronos: Learning the language of time series.
\newblock \emph{Transactions on Machine Learning Research}, 2024.
\newblock ISSN 2835-8856.
\newblock URL \url{https://openreview.net/forum?id=gerNCVqqtR}.

\bibitem[Awais et~al.(2025)Awais, Naseer, Khan, Anwer, Cholakkal, Shah, Yang, and Khan]{awais2025foundation}
Muhammad Awais, Muzammal Naseer, Salman Khan, Rao~Muhammad Anwer, Hisham Cholakkal, Mubarak Shah, Ming-Hsuan Yang, and Fahad~Shahbaz Khan.
\newblock Foundation models defining a new era in vision: a survey and outlook.
\newblock \emph{IEEE Transactions on Pattern Analysis and Machine Intelligence}, 2025.

\bibitem[Bronstein et~al.(2021)Bronstein, Bruna, Cohen, and Veli{\v{c}}kovi{\'c}]{bronstein2021GDL}
Michael~M Bronstein, Joan Bruna, Taco Cohen, and Petar Veli{\v{c}}kovi{\'c}.
\newblock Geometric deep learning: Grids, groups, graphs, geodesics, and gauges.
\newblock \emph{arXiv preprint arXiv:2104.13478}, 2021.

\bibitem[Cohen et~al.(2025)Cohen, Khwaja, Doubli, Lemaachi, Lettieri, Masson, Miccinilli, Ram{\'e}, Ren, Rostamizadeh, et~al.]{cohen2025toto}
Ben Cohen, Emaad Khwaja, Youssef Doubli, Salahidine Lemaachi, Chris Lettieri, Charles Masson, Hugo Miccinilli, Elise Ram{\'e}, Qiqi Ren, Afshin Rostamizadeh, et~al.
\newblock This time is different: An observability perspective on time series foundation models.
\newblock \emph{arXiv preprint arXiv:2505.14766}, 2025.

\bibitem[Das et~al.(2024)Das, Kong, Sen, and Zhou]{das2024timesFM}
Abhimanyu Das, Weihao Kong, Rajat Sen, and Yichen Zhou.
\newblock A decoder-only foundation model for time-series forecasting.
\newblock In \emph{Forty-first International Conference on Machine Learning}, 2024.

\bibitem[Godahewa et~al.(2021)Godahewa, Bergmeir, Webb, Hyndman, and Montero-Manso]{godahewa2021monash}
Rakshitha Godahewa, Christoph Bergmeir, Geoffrey~I. Webb, Rob~J. Hyndman, and Pablo Montero-Manso.
\newblock Monash time series forecasting archive.
\newblock In \emph{Neural Information Processing Systems Track on Datasets and Benchmarks}, 2021.

\bibitem[Goswami et~al.(2024)Goswami, Szafer, Choudhry, Cai, Li, and Dubrawski]{goswami2024moment}
Mononito Goswami, Konrad Szafer, Arjun Choudhry, Yifu Cai, Shuo Li, and Artur Dubrawski.
\newblock {MOMENT}: A family of open time-series foundation models.
\newblock In \emph{Forty-first International Conference on Machine Learning}, 2024.
\newblock URL \url{https://openreview.net/forum?id=FVvf69a5rx}.

\bibitem[Huh et~al.(2024)Huh, Cheung, Wang, and Isola]{huh2024PRH}
Minyoung Huh, Brian Cheung, Tongzhou Wang, and Phillip Isola.
\newblock Position: The platonic representation hypothesis.
\newblock In \emph{Forty-first International Conference on Machine Learning}, 2024.

\bibitem[Liu et~al.(2024)Liu, Zhang, Li, Huang, Wang, and Long]{liu2024timer}
Yong Liu, Haoran Zhang, Chenyu Li, Xiangdong Huang, Jianmin Wang, and Mingsheng Long.
\newblock Timer: generative pre-trained transformers are large time series models.
\newblock In \emph{Proceedings of the 41st International Conference on Machine Learning}, ICML'24. JMLR.org, 2024.

\bibitem[Masserano et~al.(2025)Masserano, Ansari, Han, Zhang, Faloutsos, Mahoney, Wilson, Park, Rangapuram, Maddix, and Wang]{masserano2025wavetoken}
Luca Masserano, Abdul~Fatir Ansari, Boran Han, Xiyuan Zhang, Christos Faloutsos, Michael~W. Mahoney, Andrew~Gordon Wilson, Youngsuk Park, Syama~Sundar Rangapuram, Danielle~C. Maddix, and Bernie Wang.
\newblock Enhancing foundation models for time series forecasting via wavelet-based tokenization.
\newblock In \emph{Forty-second International Conference on Machine Learning}, 2025.
\newblock URL \url{https://openreview.net/forum?id=B6WalMoQJW}.

\bibitem[Qi et~al.(2025)Qi, Hu, Lei, Zhang, Shi, Huang, Chen, Lin, and Shen]{qi2025timeHF}
Yongzhi Qi, Hao Hu, Dazhou Lei, Jianshen Zhang, Zhengxin Shi, Yulin Huang, Zhengyu Chen, Xiaoming Lin, and Zuo-Jun~Max Shen.
\newblock Timehf: Billion-scale time series models guided by human feedback.
\newblock \emph{arXiv preprint arXiv:2501.15942}, 2025.

\bibitem[Woo et~al.(2024)Woo, Liu, Kumar, Xiong, Savarese, and Sahoo]{woo2024uni2ts}
Gerald Woo, Chenghao Liu, Akshat Kumar, Caiming Xiong, Silvio Savarese, and Doyen Sahoo.
\newblock Unified training of universal time series forecasting transformers.
\newblock In \emph{Proceedings of the 41st International Conference on Machine Learning}, ICML'24. JMLR.org, 2024.

\bibitem[Xu et~al.(2025)Xu, Gupta, Cheng, Shen, Shen, Talwalkar, and Khodak]{xu2025specialized}
Zongzhe Xu, Ritvik Gupta, Wenduo Cheng, Alexander Shen, Junhong Shen, Ameet Talwalkar, and Mikhail Khodak.
\newblock Specialized foundation models struggle to beat supervised baselines.
\newblock In \emph{The Thirteenth International Conference on Learning Representations}, 2025.
\newblock URL \url{https://openreview.net/forum?id=JYTQ6ELUVO}.

\bibitem[Zhou et~al.(2021)Zhou, Zhang, Peng, Zhang, Li, Xiong, and Zhang]{zhou2021informer}
Haoyi Zhou, Shanghang Zhang, Jieqi Peng, Shuai Zhang, Jianxin Li, Hui Xiong, and Wancai Zhang.
\newblock Informer: Beyond efficient transformer for long sequence time-series forecasting.
\newblock In \emph{Proceedings of the AAAI conference on artificial intelligence}, volume~35, pages 11106--11115, 2021.

\end{thebibliography}










\end{document}